\documentclass{article}

\usepackage{PRIMEarxiv}
\usepackage{multirow}
\usepackage{threeparttable}

\usepackage[utf8]{inputenc} 
\usepackage[T1]{fontenc}    
\usepackage[colorlinks=true, linkcolor=blue, citecolor=blue, urlcolor=blue]{hyperref}       
\usepackage{url}            
\usepackage{booktabs}       
\usepackage{amsfonts}       
\usepackage{nicefrac}       
\usepackage{microtype}      
\usepackage{lipsum}
\usepackage{fancyhdr}       
\usepackage{graphicx}       
\graphicspath{{media/}}     
\usepackage{listings}
\usepackage{xcolor}

\definecolor{codegray}{gray}{0.95}
\definecolor{codepurple}{rgb}{0.58,0,0.82}
\definecolor{codeblue}{rgb}{0.25,0.5,0.75}
\definecolor{codegreen}{rgb}{0,0.6,0}

\lstdefinestyle{pythonstyle}{
    language=Python,
    backgroundcolor=\color{codegray},
    basicstyle=\ttfamily\small,
    keywordstyle=\color{codeblue}\bfseries,
    commentstyle=\color{codegreen}\itshape,
    stringstyle=\color{codepurple},
    numbers=left,
    numberstyle=\tiny,
    stepnumber=1,
    numbersep=8pt,
    showstringspaces=false,
    breaklines=true,
    tabsize=4,
    frame=single,
    captionpos=b
}

\title{GEAK: Introducing Triton Kernel AI Agent \& Evaluation Benchmarks
}

\author{
  Jianghui Wang\footnotemark[1]   \\
   \And
   Vinay Joshi\footnotemark[1]  \\
   \And
   Saptarshi Majumder\\
  \And
  Xu Chao\\
  \And
  Bin Ding \\
  \And
  Ziqiong Liu \\
  \And
  Pratik Prabhanjan Brahma \\
  \And
  Dong Li \\
  \And
  Zicheng Liu \\
  \And
  Emad Barsoum \\
  \\
  \AND
  \textit{Advanced Micro Devices, Inc. (AMD)}
}

\begin{document}
\maketitle

\footnotetext[1]{Primary authors}

\begin{abstract}

The demand for AI-generated GPU kernels is rapidly growing, influenced by the need for scalable, hardware-optimized solutions in both industry and academia. As deep learning workloads grow in complexity and diversity, it is imperative to automate low-level kernel development to meet performance and productivity demands. Major cloud providers, semiconductor companies, and research institutions are now investing heavily in AI-driven code generation for GPUs, aiming to reduce manual optimization efforts while achieving near-expert performance on hardware like AMD~Instinct™~MI300X. The Triton language, a Python-based DSL for GPU programming, has emerged as a popular target for such AI-generated kernels due to its balance of performance and ease-of-coding. In this work, we present an evaluation suite for Triton-based GPU kernels and GEAK (Generating Efficient AI-centric GPU Kernels)—a framework that leverages cutting-edge LLMs to generate performant Triton code specifically for AMD GPUs, including the AMD Instinct™ MI300X and MI250. GEAK leverages inference-time compute scaling to produce Triton-based GPU kernels using a reasoning loop adapted from Reflexion-style feedback mechanisms. On two evaluation benchmarks, GEAK significantly outperformed the baselines of directly prompting frontier LLMs as well as Reflexion-based generation pipelines by achieving correctness up to $63$\% and execution speed up of up to 2.59X. These results highlight the promise of GEAK-like agentic code generation for accelerating the adoption of diverse hardware platforms and democratizing access to expert-level kernel performance.

\end{abstract}

\keywords{Triton language \and GPU kernel generation \and Automated Code Generation \and Inference time scaling}

\section{Introduction}

\begin{figure}
    \centering
    \includegraphics[width=0.95\linewidth]{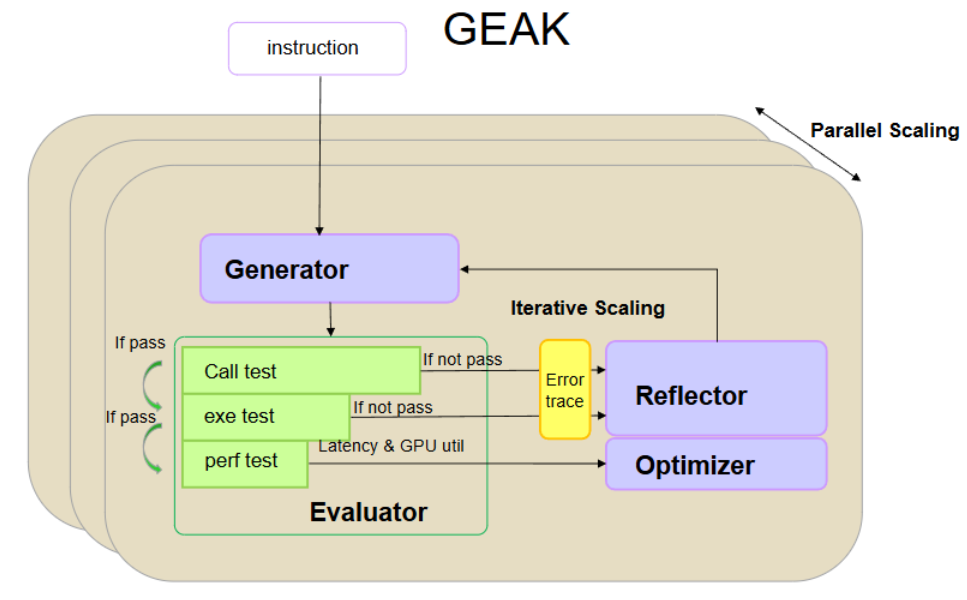}
    \caption{Illustration of GEAK pipeline.}
    \label{fig:geak-pipeline}
\end{figure}

As AI workloads scale in both complexity and hardware diversity, there is a growing demand for intelligent systems that can generate high-performance GPU kernels without manual tuning. This demand is especially critical in environments where tight coupling between software and hardware efficiency determines practical feasibility—ranging from hyperscaler data centers used for serving frontier LLMs to academic HPC clusters. Recent advances in large language models (LLMs) have shown promising capabilities in code generation tasks, but achieving correct and efficient GPU code remains an open challenge. AI-assisted GPU kernel development, particularly for emerging hardware like AMD Instinct™ MI300X GPUs, can substantially improve productivity and deployment speed.

To this end, we introduce GEAK (Generating Efficient AI-centric GPU Kernels), AMD’s new agentic framework for automatic Triton kernel generation targeting AMD Instinct™ GPUs. GEAK is built to push the frontier of AI-assisted code generation by combining state-of-the-art LLMs with a structured reasoning and feedback loop. Alongside GEAK, we release two benchmark suites for evaluating AI-generated Triton kernels, enabling measurement of both execution correctness and runtime performance of generated kernels.

\textbf{Key Contributions:}
\begin{enumerate}
    \item We introduce \textbf{GEAK framework}, a modular, agent-based system that uses inference-time compute scaling with frontier LLMs to generate Triton GPU kernels from minimal task descriptions. GEAK incorporates multiple agents—generation, evaluation, reflection, and optimization—to iteratively refine code quality and execution performance.
    \item We establish two benchmark suites for evaluating Triton kernel generation. The \textbf{TritonBench-revised} is a revised subset of 184 kernels adapted from TritonBench-G\cite{li2025tritonbenchbenchmarkinglargelanguage} with stricter testing harnesses. In addition, we release the \textbf{ROCm Triton Benchmark}, a new suite of 30 real-world kernels sampled from open-source AMD ROCm repositories.
    \item We demonstrate that GEAK significantly outperforms direct prompting of state-of-the-art LLMs, achieving correct kernel generation rate up to $54.89$\% on TritonBench-revised and $63.33$\% on the ROCm Triton benchmark—compared to less than $15$\% when directly prompting strong LLMs without agentic feedback.
    \item We also present a detailed study on one of the performant kernels from ROCm Triton Benchmark (please see Appendix \ref{appendix:case-study}).
    \item We open source our GEAK agent implementation (\href{https://github.com/AMD-AIG-AIMA/GEAK-agent}{github link}) and GEAK evaluation framework (\href{https://github.com/AMD-AIG-AIMA/GEAK-eval}{github link}).
\end{enumerate}

Furthermore, GEAK-generated kernels demonstrate an average speedup of up to $2.59$× over their reference counterparts on the TritonBench-revised set. 

\section{Related Work}
\label{sec:headings}
\paragraph{Benchmarks}
As program synthesis research advances, the need for rigorous and multidimensional benchmarks has become increasingly important. Most existing coding benchmarks concentrate on functional correctness, often assessed through manually crafted test cases and isolated execution environments. For instance, HumanEval provides expert-written programming tasks paired with curated test suites, while MBPP\cite{austin2021programsynthesislargelanguage} collects problems through crowdsourcing. To improve coverage and scalability, more recent approaches have turned to automated test generation\cite{liu2023codegeneratedchatgptreally}, enabling broader application in areas such as software engineering and developer assistance\cite{jimenez2024swebenchlanguagemodelsresolve}.

In addition to correctness, performance profiling is gaining traction as a crucial axis in benchmarking (\cite{shypula2024learningperformanceimprovingcodeedits};\cite{liu2023codegeneratedchatgptreally};\cite{huang2025effibenchbenchmarkingefficiencyautomatically};\cite{qiu2025efficientllmgeneratedcoderigorous}). However, most existing frameworks emphasize competition-style tasks or sequential execution and provide limited support for parallel or hardware-specific evaluation. While a few benchmarks address parallel CPU programming (\cite{Nichols_2024};\cite{chaturvedi2024hpccoderv2studyingcodellms}), GPU-targeted benchmarks remain scarce, despite the growing importance of hardware efficiency in deploying large-scale deep learning models.

To address this gap, benchmarks like KernelBench~\cite{ouyang2025kernelbenchllmswriteefficient} and TritonBench~\cite{li2025tritonbenchbenchmarkinglargelanguage} have emerged, offering domain-specific assessments of LLM performance on GPU kernel generation. KernelBench~\cite{ouyang2025kernelbenchllmswriteefficient} focuses on correctness and runtime efficiency across diverse workloads using metrics like "fast\_p", while TritonBench~\cite{li2025tritonbenchbenchmarkinglargelanguage} highlights the challenges LLMs face in generating performant code for the Triton DSL. These benchmarks reveal a substantial gap between general-purpose code generation and the demands of high-performance, hardware-aware synthesis, underscoring the need for specialized, compute-oriented evaluation frameworks.

\paragraph{LLMs for Code Generation}

More recent work has explored the use of large language models for code generation, but often in general-purpose contexts with limited success on performance-critical low level code, such as DeepSeek-Coder\cite{guo2024deepseekcoderlargelanguagemodel} and Qwen-Coder\cite{hui2024qwen25codertechnicalreport}, which have achieved strong performance on broad coding benchmarks.

There are also LLMs specifically trained for the kernel generation task, such as KernelLLM. However, its capability is limited to translating PyTorch code into Triton kernels, restricting its general applicability. In contrast, our agent can generate Triton code based on both natural language instructions and reference code, allowing for more flexible and context-aware code synthesis. Furthermore, specialized models like KernelLLM often lack the broader reasoning and problem-solving capabilities exhibited by larger state-of-the-art LLMs such as GPT-4.1 and Claude 4, which are critical for tackling complex optimization tasks involving multiple constraints and trade-offs.

\paragraph{Inference-Time Compute Scaling}
Inference-time compute scaling refers to the strategy of allocating increased computational resources—such as longer context windows, multiple reasoning passes, or parallel generations — during the inference phase of machine learning models to boost performance on complex tasks.
Notably, this approach enables significant improvements in large language model (LLM) performance without requiring additional training.
For example, the Chain-of-Thought~\cite{shinn2023reflexionlanguageagentsverbal} prompting technique enhances reasoning by encouraging models to answer questions in a step-by-step manner. Similarly, the Reflexion~\cite{shinn2023reflexionlanguageagentsverbal} framework improves agent performance by integrating verbalized feedback into an episodic memory buffer, which guides more informed decision-making in future attempts.
Our proposed GEAK agent also leverages inference-time compute scaling to improve code generation and optimization quality. In addition, we incorporate a Reflexion-style module for debugging, enabling the agent to iteratively refine its outputs based on past execution feedback.

\section{Benchmarks}
\label{sec:others}

We experiment with Triton kernel evaluation benchmarks, consisting of a modified set of 184 kernels originally sourced from TritonBench-G and an additional set of 30 kernels from various open-sourced ROCm repositories.

\subsection{TritonBench-revised Benchmark}

The Triton kernels and corresponding evaluation codebase were borrowed and adapted from TritonBench but the following enhancements were made: 

37 out of the 184 kernels in TritonBench-G~\cite{li2025tritonbenchbenchmarkinglargelanguage} had errors while running on AMD GPUs (such as Shared memory errors, invalid HIP arguments, ModuleNotFound). We fixed these kernel to be AMD GPU compliant.

For several kernels, the original TritonBench repository had missing call functions for which execution accuracy script was effectively comparing empty strings as output. This was corrected by introducing the following:
\begin{itemize}
    \item Adding call to the test functions in the kernels missing it.
    \item Developing tolerance based tensor comparison of output tensors produced by test functions instead of STDOUT string comparisons.
    \item Ensuring consistent seed for random tensor generations used in producing inputs in unit tests.
\end{itemize}

For some kernels, unit tests were written inconsistently that produced unexpected outputs. We have fixed these kernels to have consistent unit tests.

\subsection{ROCm Triton Benchmark}

The following kernels have been written by Triton expert engineers and released publicly to help enable the ecosystem around running AI workloads efficiently on AMD GPUs. We also took assistance from frontier LLMs to refactor these, as well as their corresponding unit tests, in the same format as the TritonBench-revised benchmark. This enables us to compare the accuracy and efficiency of our agent in a consistent manner.

\begin{table}
 \caption{List of kernels and their sources in the RoCM repositories Triton benchmark}
  \centering
  \begin{tabular}{ll}
    \toprule
    \cmidrule(r){1-2}
    ROCm Kernel     & Repository link  \\
    \midrule
    1. test\_tma\_store\_gemm & \multirow{10}{*}{ROCm/triton }\\
    2. moe-gemm    &\\
    3. layernorm   &\\
    4. triton\_multreduce\_matmul\_kernel.py &\\
    5. multreduce\_matmul\_dot\_kernel.py    &\\
    6. gemm       &\\
    7. test\_block\_pointer\_matmul  &\\
    8. test\_block\_copy     &\\
    9. test\_add\_kernel   & \\
    10. softmax &\\
    \midrule
    1. rmsnorm\_fwd   &  \multirow{5}{*}{ROCm/aiter}\\
    2. rmsnorm\_bwd   & \\
    3. test\_chained\_dot\_fp8 &\\
    4. test\_matmul\_MXFP & \\
    5. test\_gemm\_fusion & \\
    \midrule
    1. test\_triton\_swizzle2d  &  \multirow{4}{*}{ROCm/aotriton}\\
    2. test\_triton\_sort &\\
    3. test\_triton\_flip &\\
    4. test\_reverse\_range &\\
    \midrule
    1. test\_load\_reduce.py & \multirow{2}{*}{ROCm/vllm} \\
    2. test\_chained\_matmul &\\
    \midrule
    1. test\_random\_int & \multirow{5}{*}{ROCm/pytorch}\\
    2. test\_randn    & \\
    3. test\_kernel\_sub &\\
    4. test\_kernel\_dot &\\
    5. test\_batched\_vecmat &\\
    \bottomrule
    1.test\_flashattention\_fwd & \multirow{2}{*}{ROCm/xformers}\\
    2.test\_iv\_dependent\_matmul &\\
    \midrule
    1.test\_gemm\_no\_scf & ROCm/bitsandbytes \\
    \midrule
    1.test\_cast\_matmul &ROCm/TransformerEngine \\
    \bottomrule
  \end{tabular}
  \label{tab:table}
\end{table}

\subsection{Evaluation Metrics}
We use the same metrics as reported in TritonBench \cite{li2025tritonbenchbenchmarkinglargelanguage}: 

\begin{itemize}
    \item \textbf{Call Accuracy:} Fraction of AI-generated kernels that can compile and run without any errors. 
    \item \textbf{Execution Accuracy:} Percentage of AI-generated kernels that satisfy all unit tests.

    Execution accuracy is computed for kernels that successfully compile.
    
    \item \textbf{Speedup:} Relative execution time improvement of AI-generated kernels over reference ground truth kernels. We define it as: expectation of the ratio of median reference kernel latency and generated kernel latency over all unit tests.

    Speedup is only computed for kernels that satisfy all unit tests.

\end{itemize}

\section{GEAK}
\subsection{Pipeline}
\label{sec:geak-pipeline}

As shown in Figure \ref{fig:geak-pipeline}, the agentic AI system comprises four core modules: 1) Generator, 2) Reflector, 3) Evaluator, and 4) Optimizer. The Generator produces code based on user query and contextual information. The Evaluator follows a cascaded design: it first performs a functionality test to verify correctness. If the code fails this test, the corresponding error trace is fed back to the Reflector for further analysis. If the code passes, the Evaluator proceeds to assess its performance, including latency and memory efficiency. If the generated code fails to execute correctly, the Reflector analyzes both the code and the resulting error trace to identify potential issues. Finally, the Optimizer takes as input the functionally correct code and formulates strategies to enhance its performance with respect to latency and computational efficiency. 
To improve the performance of our agent, we employ some composable and configurable techniques as mentioned in below.

\subsection{Modules}
\paragraph{1- shot prompting}

To enable 1-shot prompting, the most similar Triton code from existing datasets is used in the prompt. The datasets used for 1-shot prompt do not overlap with the benchmark. We observed that 1-shot sample retrieval from datasets is most effective with code similarity rather than instruction similarity. 

\paragraph{Knowledge Injection} 

We enhance the prompt with domain-specific knowledge on writing efficient Triton kernels, including detailed hardware specifications. This incorporation of low-level optimization principles significantly improves the accuracy and quality of LLM-generated code. 

\paragraph{Reflexion}

To enhance the system's self-correcting capabilities, we leverage a Reflexion\cite{shinn2023reflexionlanguageagentsverbal} module that supports introspective debugging and iterative refinement. When an LLM-generated kernel does not pass the functionality test, the resulting error trace is provided as feedback to the reflector for further analysis and correction. The agent is tasked with analyzing the cause of failure and proposing a corresponding solution. We scale the number of iterations to improve kernel generation. 

\paragraph{LLM selection for kernel agent}  

We have explored the use of multiple LLMs like GPT-4.1, O1 and Gemini 2.5 Pro. We observed significant variation in the outputs produced by different models, indicating that the capability of the underlying LLM can substantially influence the results. 

\paragraph{LLM as Optimizer} 

The Optimizer LLM is tasked with identifying potential optimization directions based on previous code generations and their corresponding performance metrics. These historical records, which include the generated code and associated performance results, are sorted in ascending order of performance. This structured presentation helps guide the LLM toward proposing more effective optimization strategies. \cite{yang2024largelanguagemodelsoptimizers}

\paragraph{Debugging trap} 

When LLM’s generated code has bugs, the error trace is provided to the Reflector for correction. However, we've observed that sometimes code can undergo several reflection cycles while still being plagued by the same bug, which is what we refer to as the debugging trap. To prevent the agent from getting stuck in a debugging trap, we impose a limit on the number of debugging attempts per code snippet using a parameter max\_perf\_debug\_num. If the code continues to fail after reaching this threshold, the agent is required to discard the current approach and generate a new strategy along with fresh code. 

\paragraph{Parallel Scaling}  

To ensure reliable kernel generation, we run multiple instances of GEAK multiple times in parallel and independently. To introduce diversity in the generated code, the LLM output is sampled with the parameter temperature set to 1. Our experiments show that such diversity in generation can yield correct and faster kernels otherwise underexplored. Moreover, further experiments indicate that combining sequential and parallel scaling yields additional performance improvements. To obtain unbiased estimates of accuracy, we estimate it using pass@k metric.

\section{Experiments}

\subsection{Main Results}
We first establish our baseline by directly prompting frontier LLMs, such as GPT-4.1, Gemini 2.5 Pro, and Claude 3.7 Sonnet, to produce the kernels. As shown in Table\ref{tab:baselines}, across all models, direct prompting yields low execution accuracy and suboptimal performance, indicating that even state-of-the-art LLMs struggle to produce correct and efficient low-level code without further guidance.

Introducing one-shot prompting—where a single example retrieved from TritonBench trainset is provided—consistently improves both call accuracy and execution accuracy compared to zero-shot prompting. In most cases, one-shot prompting also leads to higher speedup, suggesting that the additional context helps LLMs generate more efficient kernels. However, a notable exception is observed with Gemini 2.5 Pro, where one-shot prompting results in lower speedup than direct prompting, despite improvements in accuracy.

Table\ref{tab:baselines} shows the direct prompting results on the ROCm benchmark. Direct prompting with GPT-4.1 fails to generate any valid kernels, underscoring the challenge of targeting non-CUDA platforms. In contrast, Gemini 2.5 Pro achieves better results than GPT-4.1 on both the TritonBench-revised and ROCm benchmarks, though the overall performance remains limited.

As shown in Table\ref{tab:geak_on_tritonbench} and Table\ref{tab:geak_on_rocm_mi300}, compared to direct LLM prompting, our system GEAK delivers significantly stronger results. On MI300, GEAK achieves 54.89\% execution accuracy and a 2.59× speedup on TritonBench-revised Benchmark, and 63.33\% execution accuracy and 0.92x speedup on ROCm Benchmark, demonstrating the effectiveness of our agent framework. We also present a case study in Appendix \ref{appendix:case-study} on $triton\_test\_flip.py$ kernel analyzing potential strategies for performance gain.

\begin{table}
 \caption{Baseline Results}
  \centering
  \begin{threeparttable}
  \begin{tabular}{llll}
    \toprule
    Model     & Call Accuracy (\%)     & Exec Accuracy {\%} & Speedup \\
    \midrule
    \multicolumn{4}{c}{TritonBench-modified benchmark}                   \\
    \midrule
    GPT4.1 & 14.67 / 19.02  &  8.70 / 14.13  & 0.52 / 0.53    \\
    GPT4o     &  10.87 / 14.13  & 7.07 / 9.24 & 0.51 / 0.53      \\
    Gemini2.5pro     & 20.65 / 21.74       & 14.13 / 16.85 & 1.33 / 0.96  \\
    Claude 3.7  & 11.41 / 20.11 & 7.07 / 15.22 & 0.61 / 0.96 \\
    \midrule
    \multicolumn{4}{c}{ROCm benchmark (0-shot results only) } \\
    \midrule
    GPT 4.1 & 0 & 0 & 0 \\
    Gemini-2.5 Pro & 40 & 16.66 & 0.91 \\
    \bottomrule
  \end{tabular}
  \label{tab:baselines}
  \begin{tablenotes}
  \small
\item Note:  numbers to the left of ‘/’ indicate 0-shot results and to the right are 1-shot prompting results.
\end{tablenotes}
  \end{threeparttable}
\end{table}
\begin{table}
 \caption{GEAK on TritonBench-Modified Benchmark }
  \centering
  \begin{threeparttable}
  \begin{tabular}{lllll}
    \toprule
    Difficulty level & Correctly generated kernels & Total kernels in dataset & Exec Accuracy (\%) & Average Speedup  \\
    \midrule
    1 & 2 / 3 & 3 & 66.67 / 100.0 & 1.24 / 1.16 \\
    2 & 23 / 22 & 27 & 85.19 / 81.48 & 4.78 / 1.69 \\
    3 & 39 / 41 & 65 & 60.00 / 63.08 & 1.57 / 3.02 \\
    4 & 32 / 34 & 84 & 38.10 / 40.48 & 2.24 / 2.86 \\
    5 & 1 / 1 & 5 & 20.00 / 20.00 & 0.14 / 0.61 \\
    \midrule
    overall & 97 / 101 & 184 & 52.72 / 54.89 & 2.42 / 2.59 \\
    \bottomrule
  \end{tabular}
  \label{tab:geak_on_tritonbench}
  \begin{tablenotes}
  \small
\item Note:  numbers to the left of ‘/’ indicate results on MI250 and to the right are results on MI300.
\end{tablenotes}
  \end{threeparttable}
\end{table}
\begin{table}
 \caption{GEAK on ROCm Benchmark}
  \centering
  \begin{tabular}{lllll}
    \toprule
    Difficulty level & Correctly generated kernels & Total kernels in dataset & Exec Accuracy (\%) & Average Speedup  \\
    \midrule
    overall & 19 & 30 & 63.33 & 0.92 \\
    \bottomrule
  \end{tabular}
  \label{tab:geak_on_rocm_mi300}
\end{table}

\subsection{Ablation Experiments}

\subsubsection{Sequential Scaling}
We investigate the impact of increasing inference-time compute on accuracy and performance, focusing on the sequential scaling dimension.

The experimental results in Table\ref{tab:sequential_scaling_mi250} indicate that increasing the number of iterations leads to a steady improvement in both the accuracy and performance of GEAK. As the number of iterations increases, we observe a monotonic improvement in both call and execution accuracy, culminating at iter19 with 63.04\% call accuracy and 44.02\% execution accuracy—more than 3× higher than the baseline.

In terms of performance, speedup improves significantly in the early iterations, with iter1 already reaching 1.89×, the highest across all settings. This sharp gain is likely due to early corrections of suboptimal code generations. Although the speedup fluctuates slightly in later iterations, it consistently stays above 1.4× from iter7 onwards, suggesting that continued refinement contributes to runtime efficiency as well.

Notably, iteration count beyond 10 still yields meaningful gains: call accuracy increases from 56.52\% (iter9) to 63.04\% (iter19), and execution accuracy from 40.76\% to 44.02\%. These results validate the effectiveness of iterative refinement in improving both correctness and efficiency.

Overall, the analysis confirms that sequential compute budget is a powerful scaling axis for GEAK, enabling it to progressively refine outputs with improved correctness and performance.

\begin{table}
 \caption{Sequential Scaling of GEAK on TritonBench-Modified Benchmark on MI250}
  \centering
  \begin{tabular}{llll}
    \toprule
    \#iterations & call acc(\%) & exec acc(\%) & speedup  \\
    \midrule
    iter0 & 21.2 & 13.04 & 1.02 \\
    iter1 & 33.15 & 26.63 & \textbf{1.89} \\
    iter2 & 41.85 & 30.98 & 1.69 \\
    iter3 & 47.28 & 34.24 & 1.39 \\
    iter4 & 50.54 & 37.5 & 1.39  \\
    iter5 & 57.61 & 37.5 & 1.63 \\
    iter6 & 50.04 & 37.5 & 1.31 \\
    iter7 & 54.39 & 39.13 & 1.48 \\
    iter8 & 62.5 & 40.22 & 1.67 \\
    iter9 & 56.52 & 40.76 & 1.45 \\
    iter10 & 62.5 & 40.22 & 1.75 \\
    iter11 & 55.43 & 40.76 & 1.45 \\
    iter12 & 57.07 & 42.39 & 1.16 \\
    iter13 & 59.78 & 41.85 & 1.75 \\
    iter14 & 61.41 & 40.76 & 1.76 \\
    iter15 & 57.07 & 43.48 & 1.65 \\
    iter16 & 60.33 & 42.93 & 1.57 \\
    iter17 & 60.87 & 43.48 & 1.85 \\
    iter18 & 61.96 & 43.48 & 1.74 \\
    iter19 & \textbf{63.04} & \textbf{44.02} & 1.55 \\
   
    \bottomrule
  \end{tabular}
  \label{tab:sequential_scaling_mi250}
\end{table}

\subsubsection{Parallel Scaling}
We demonstrate how increasing inference-time compute through parallel scaling improves both accuracy and performance. Specifically, we fix the number of sequential iterations to 10 and explore how varying the number of parallel runs during inference affects the results. Experiments are conducted on both MI250 and MI300 GPUs to ensure the findings generalize across hardware generations.

\begin{figure}
  \centering
  \includegraphics[width=0.85\linewidth]{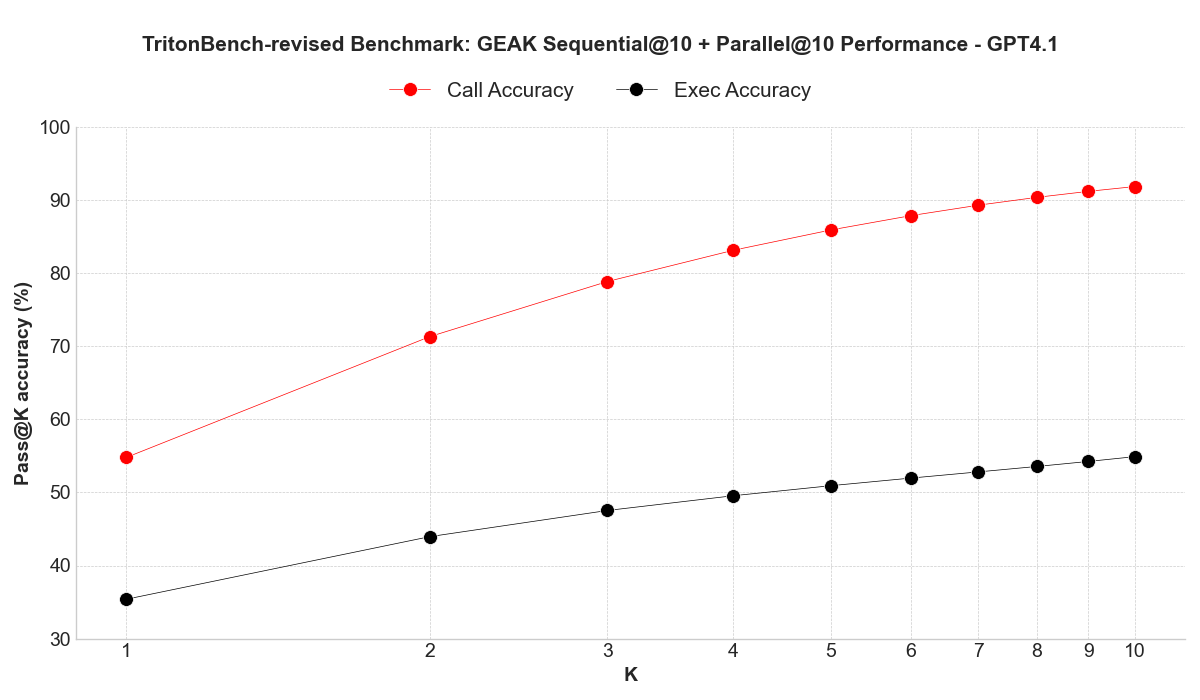}
  \caption{Inference time compute scaling study on TritonBench-revised benchmark.}
  \label{fig:fig1}
\end{figure}

\begin{figure}
  \centering
  \includegraphics[width=0.85\linewidth]{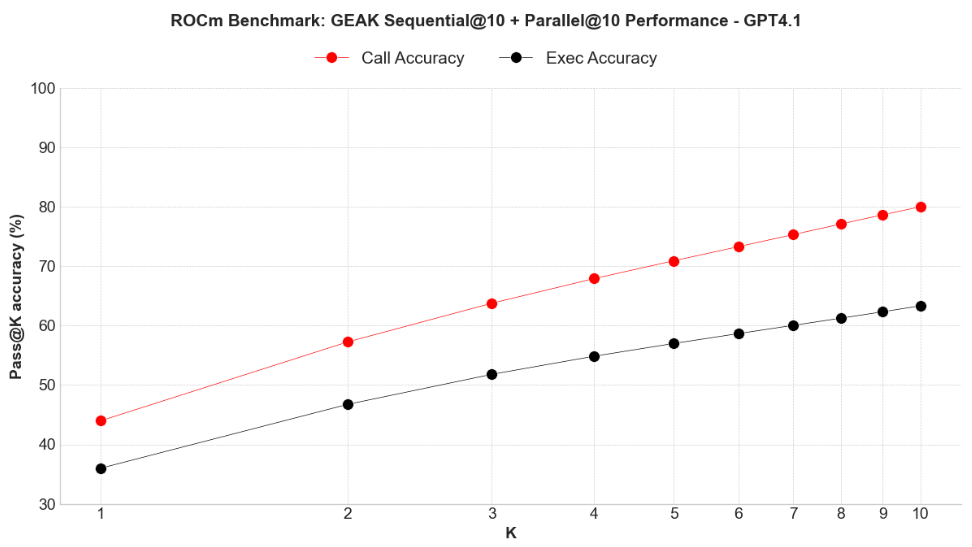}
  \caption{Inference time compute scaling study on ROCm benchmark.}
  \label{fig:fig2}
\end{figure}

Figure\ref{fig:fig1} and Figure\ref{fig:fig2} show inference time scaling laws on ROCm and TritonBench-revised benchmarks, respectively. Both call and execution accuracy scale almost log-linearly with respect to the number of parallel runs,

These findings underscore the complementary role of parallel scaling alongside sequential refinement as two orthogonal axes to boost inference quality. Together, they provide a flexible mechanism for tuning the accuracy-performance tradeoff under a given compute budget.

The parallel scaling analysis reveals several critical insights for the GEAK framework's design and implementation strategy. First, parallel scaling provides complementary benefits to sequential refinement, enabling the framework to achieve higher accuracy without proportionally increasing latency. This orthogonal improvement axis allows GEAK to optimize both accuracy and efficiency simultaneously.

The consistent scaling patterns across different hardware platforms demonstrate that GEAK's parallel scaling mechanisms are robust and portable. The framework can leverage available computational resources effectively regardless of the underlying GPU architecture, making it suitable for diverse deployment environments.

The substantial improvements in Call Accuracy relative to Execution Accuracy indicate that parallel scaling particularly benefits high-level reasoning and decision-making processes. This suggests that GEAK's parallel mechanisms are especially valuable for complex inference tasks that require sophisticated logical reasoning rather than simple execution steps.

Finally, the scaling characteristics support GEAK's flexible architecture design, where parallel compute resources can be dynamically allocated based on task complexity and accuracy requirements. This adaptability ensures that the framework can optimize performance across different use cases while maintaining computational efficiency.

\begin{table}
 \caption{TritonBench-Modified Benchmark: GEAK Sequential@10 + Parallel@10 on MI250 - GPT4.1}
  \centering
  \begin{tabular}{lll}
    \toprule
    Pass@K & call acc(\%) & exec acc(\%) \\
       \midrule
    1 & 56.52 & 40.76 \\
    2 & 66.30 & 45.65 \\
    3 & 72.83 & 48.91 \\
    4 & 78.80 & 50.0 \\
    5 & 81.52 & 50.0 \\
    6 & 82.61 & 49.46 \\
    7 & 85.87 & 52.17 \\
    8 & 86.41 & 53.26 \\
    9 & 86.41 & 53.26 \\
   
    \bottomrule
  \end{tabular}
  \label{tab:parallel_scaling_tritonbench_mi250}
\end{table}
\begin{table}
 \caption{TritonBench-Modified Benchmark: GEAK Sequential@10 + Parallel@10 on MI300 - GPT4.1}
  \centering
  \begin{tabular}{lll}
    \toprule
    Pass@K & call acc(\%) & exec acc(\%) \\
       \midrule
    1 & 54.78 & 35.38 \\
    2 & 71.31 & 43.95 \\
    3 & 78.85 & 47.53 \\
    4 & 83.13 & 49.55 \\
    5 & 85.92 & 50.92 \\
    6 & 87.87 & 51.96 \\
    7 & 89.30 & 52.81 \\
    8 & 90.37 & 53.55 \\
    9 & 91.20 & 54.24 \\
    10 & 91.85 & 54.89 \\
   
    \bottomrule
  \end{tabular}
  \label{tab:parallel_scaling_tritonbench_mi300}
\end{table}
\begin{table}
 \caption{ROCm Benchmark: GEAK Sequential@10 + Parallel@10 on MI300 - GPT4.1}
  \centering
  \begin{tabular}{lll}
    \toprule
    Pass@K & call acc(\%) & exec acc(\%) \\
       \midrule
   1 & 43.99 & 36.00 \\
   2 & 57.25 & 46.74 \\
   3 & 63.77 & 51.80 \\
   4 & 67.92 & 54.85 \\
   5 & 70.91 & 56.99 \\
   6 & 73.30 & 58.66 \\
   7 & 75.33 & 60.05 \\
   8 & 77.11 & 61.25 \\
   9 & 78.66 & 62.33 \\
   10 & 80.00 & 63.33 \\
   
    \bottomrule
  \end{tabular}
  \label{tab:parallel_scaling_rocm_mi300}
\end{table}

\subsubsection{Effect of different Modules}
To assess the contribution of individual components within our agent architecture, we conducted ablation experiments by evaluating performance across various combinations of modules.
Results in Table\ref{tab:modules_ablation} show that each module—knowledge injection, one-shot, and Optimizer—contributes positively to both execution accuracy and speedup, highlighting the complementary roles these components play in enhancing the agent's overall performance. 

Specifically, knowledge injection alone provides a substantial lift over the baseline, improving call accuracy from 14.67\% to 52.72\% and execution accuracy from 8.70\% to 20.11\%. This underscores the critical role of prior knowledge in enabling correct kernel generation.

Adding one-shot prompting further improves accuracy, pushing call accuracy to 54.35\% and execution accuracy to 27.17\%. However, the speedup remains under 1.0 (i.e., 0.99), suggesting that while one-shot helps with correctness, it has limited impact on runtime efficiency on its own.

The introduction of the Optimizer module shows the most significant impact on speedup. When combined with the other two modules, it boosts execution accuracy to 40.76\% and yields a notable speedup of 1.45×, the highest among all configurations. This indicates that the Optimizer not only improves the likelihood of generating correct kernels but also helps produce more efficient code in terms of latency and performance.
\begin{table}
 \caption{Modules}
  \centering
  \begin{tabular}{llllll}
    \toprule
    Knowledge Injection & 1-shot & Optimizer & call acc(\%) & exec acc(\%) & speedup \\
       \midrule
    &  &  &  14.67 & 8.70 & 0.52\\
    \checkmark & &  & 52.72 & 20.11 & 0.86 \\
    \checkmark & \checkmark & & 54.35 & 27.17 & 0.99 \\
    \checkmark & \checkmark & \checkmark & 56.52 & 40.76 & 1.45 \\
   
    \bottomrule
  \end{tabular}
  \label{tab:modules_ablation}
\end{table}

\section{Conclusion}
In this work, we presented GEAK, a modular and agent-based framework that leverages inference-time compute scaling with frontier LLMs to automatically generate efficient Triton GPU kernels from text task descriptions. By coordinating multiple specialized agents, GEAK iteratively improves both the correctness and performance of generated code without requiring additional training.

To rigorously assess kernel generation capabilities, we introduced two benchmarks: TritonBench-revised benchmark, a curated subset of TritonBench-G with improved test harnesses, and the newly constructed ROCm benchmark, which consists of real-world kernels from open-source AMD repositories. These benchmarks provide a diverse and challenging testbed for evaluating both functional correctness and runtime efficiency.

By fully open-sourcing the benchmarks and code for running the agent, we expect to engage the open-source community to accelerate the development of GPU kernels. We invite developers, researchers, and AI enthusiasts to explore the agent and benchmarks, and we hope this will foster innovation and collaboration within the AI community to come up with even better methods and final output kernels that can significantly improve the efficiency of training and inference for large-scale AI models.

\section*{Acknowledgments}
We would like to acknowledge the following folks for constructive discussions and feedback during the project - Alan Lee, Peng Sun, Vinayak Gokhale, Jason Furmanek, Sharunas Kalade, Graham Schelle, Sampsa Rikonen, Doug Lehr, Zhaoyi Li, Yonatan Dukler, Vikram Appia, Arseny Moskvichev, Stephen Youn and Steve Reinhart.

\bibliographystyle{unsrt}  
\bibliography{references}  

\begin{thebibliography}{10}

\bibitem{li2025tritonbenchbenchmarkinglargelanguage}
Jianling Li, Shangzhan Li, Zhenye Gao, Qi~Shi, Yuxuan Li, Zefan Wang, Jiacheng Huang, Haojie Wang, Jianrong Wang, Xu~Han, Zhiyuan Liu, and Maosong Sun.
\newblock Tritonbench: Benchmarking large language model capabilities for generating triton operators, 2025.

\bibitem{austin2021programsynthesislargelanguage}
Jacob Austin, Augustus Odena, Maxwell Nye, Maarten Bosma, Henryk Michalewski, David Dohan, Ellen Jiang, Carrie Cai, Michael Terry, Quoc Le, and Charles Sutton.
\newblock Program synthesis with large language models, 2021.

\bibitem{liu2023codegeneratedchatgptreally}
Jiawei Liu, Chunqiu~Steven Xia, Yuyao Wang, and Lingming Zhang.
\newblock Is your code generated by chatgpt really correct? rigorous evaluation of large language models for code generation, 2023.

\bibitem{jimenez2024swebenchlanguagemodelsresolve}
Carlos~E. Jimenez, John Yang, Alexander Wettig, Shunyu Yao, Kexin Pei, Ofir Press, and Karthik Narasimhan.
\newblock Swe-bench: Can language models resolve real-world github issues?, 2024.

\bibitem{shypula2024learningperformanceimprovingcodeedits}
Alexander Shypula, Aman Madaan, Yimeng Zeng, Uri Alon, Jacob Gardner, Milad Hashemi, Graham Neubig, Parthasarathy Ranganathan, Osbert Bastani, and Amir Yazdanbakhsh.
\newblock Learning performance-improving code edits, 2024.

\bibitem{huang2025effibenchbenchmarkingefficiencyautomatically}
Dong Huang, Yuhao Qing, Weiyi Shang, Heming Cui, and Jie~M. Zhang.
\newblock Effibench: Benchmarking the efficiency of automatically generated code, 2025.

\bibitem{qiu2025efficientllmgeneratedcoderigorous}
Ruizhong Qiu, Weiliang~Will Zeng, James Ezick, Christopher Lott, and Hanghang Tong.
\newblock How efficient is llm-generated code? a rigorous \& high-standard benchmark, 2025.

\bibitem{Nichols_2024}
Daniel Nichols, Joshua~H. Davis, Zhaojun Xie, Arjun Rajaram, and Abhinav Bhatele.
\newblock Can large language models write parallel code?
\newblock In {\em Proceedings of the 33rd International Symposium on High-Performance Parallel and Distributed Computing}, HPDC ’24, page 281–294. ACM, June 2024.

\bibitem{chaturvedi2024hpccoderv2studyingcodellms}
Aman Chaturvedi, Daniel Nichols, Siddharth Singh, and Abhinav Bhatele.
\newblock Hpc-coder-v2: Studying code llms across low-resource parallel languages, 2024.

\bibitem{ouyang2025kernelbenchllmswriteefficient}
Anne Ouyang, Simon Guo, Simran Arora, Alex~L. Zhang, William Hu, Christopher Ré, and Azalia Mirhoseini.
\newblock Kernelbench: Can llms write efficient gpu kernels?, 2025.

\bibitem{guo2024deepseekcoderlargelanguagemodel}
Daya Guo, Qihao Zhu, Dejian Yang, Zhenda Xie, Kai Dong, Wentao Zhang, Guanting Chen, Xiao Bi, Y.~Wu, Y.~K. Li, Fuli Luo, Yingfei Xiong, and Wenfeng Liang.
\newblock Deepseek-coder: When the large language model meets programming -- the rise of code intelligence, 2024.

\bibitem{hui2024qwen25codertechnicalreport}
Binyuan Hui, Jian Yang, Zeyu Cui, Jiaxi Yang, Dayiheng Liu, Lei Zhang, Tianyu Liu, Jiajun Zhang, Bowen Yu, Keming Lu, Kai Dang, Yang Fan, Yichang Zhang, An~Yang, Rui Men, Fei Huang, Bo~Zheng, Yibo Miao, Shanghaoran Quan, Yunlong Feng, Xingzhang Ren, Xuancheng Ren, Jingren Zhou, and Junyang Lin.
\newblock Qwen2.5-coder technical report, 2024.

\bibitem{shinn2023reflexionlanguageagentsverbal}
Noah Shinn, Federico Cassano, Edward Berman, Ashwin Gopinath, Karthik Narasimhan, and Shunyu Yao.
\newblock Reflexion: Language agents with verbal reinforcement learning, 2023.

\bibitem{yang2024largelanguagemodelsoptimizers}
Chengrun Yang, Xuezhi Wang, Yifeng Lu, Hanxiao Liu, Quoc~V. Le, Denny Zhou, and Xinyun Chen.
\newblock Large language models as optimizers, 2024.

\end{thebibliography}

\appendix

\section{Case study: Flip kernel}

\label{appendix:case-study}

We studied kernel $test\_triton\_flip.py$, from ROCm Triton Benchmark, which achieved $2.26$x speedup over Triton expert written kernel. The following was the triton kernel generated by GEAK:

\begin{lstlisting}[style=pythonstyle]
@triton.jit
def flip_kernel(X, Z, N: tl.constexpr, M: tl.constexpr):
    """
    Processes 2D blocks of data, flipping each block horizontally.
    Loads a (N, M) block from X, flips each row horizontally, and stores into Z.
    Parameters:
    X: pointer to flattened input tensor
    Z: pointer to flattened output tensor
    N: number of rows in the block
    M: number of columns in the block
    """
    offs_row = tl.arange(0, N)
    offs_col = tl.arange(0, M)
    rows = offs_row
    cols = offs_col
    mask_row = rows < N
    mask_col = cols < M
    mask = mask_row[:, None] & mask_col[None, :]
    flipped_cols = M - 1 - cols
    x_ptrs = X + rows[:, None] * M + flipped_cols[None, :]
    z_ptrs = Z + rows[:, None] * M + cols[None, :]
    vals = tl.load(x_ptrs, mask=mask, other=0)
    tl.store(z_ptrs, vals, mask=mask)
\end{lstlisting}

Corresponding Triton expert written kernel is below:
\begin{lstlisting}[style=pythonstyle]
@triton.jit
def flip_kernel(X, Z, N: tl.constexpr, M: tl.constexpr):
    offx = tl.arange(0, M)
    offy = tl.arange(0, N) * M
    off2d = offx[None, :] + offy[:, None]
    x = tl.load(X + off2d)
    x = tl.flip(x)
    tl.store(Z + off2d, x)
\end{lstlisting}

\subsection{Performance Analysis}

\textbf{Expert written code limitations:}
\begin{enumerate}
    \item Double Memory Access Pattern:
   \begin{lstlisting}[style=pythonstyle]
   x = tl.load(X + off2d)  # Load entire block
   x = tl.flip(x)          # Flip in registers
   tl.store(Z + off2d, x)  # Store entire block
   \end{lstlisting}
   \begin{itemize}
       \item    Loads the entire (N×M) block into registers
   \item Performs flip operation on register data
   \item Stores the entire block back to memory
   \item This creates unnecessary memory bandwidth usage
   \end{itemize}

    \item Register Pressure: 
   \begin{itemize}
   \item Must hold the entire block in registers simultaneously
   \item For large blocks, this can exceed register capacity
   \item May cause register spilling to local memory
   \end{itemize}

    \item Limited Flexibility:
    \begin{itemize}
   \item `tl.flip()` behavior may not be optimally tuned for all tensor shapes
   \item Less control over the exact memory access pattern
    \end{itemize}

\end{enumerate}

GEAK generated code advantages:
\begin{enumerate}

\item Optimized Memory Access Pattern:
   \begin{lstlisting}[style=pythonstyle]
   flipped_cols = M - 1 - cols
   x_ptrs = X + rows[:, None] * M + flipped_cols[None, :]  # Read from flipped positions
   z_ptrs = Z + rows[:, None] * M + cols[None, :]          # Write to normal positions
   vals = tl.load(x_ptrs, mask=mask, other=0)
   tl.store(z_ptrs, vals, mask=mask)
   \end{lstlisting}
   \begin{itemize}
       \item  Single pass operation: Reads from flipped source positions and writes directly to destination
   \item No intermediate storage of the entire block
   \item  More cache-friendly access pattern
   \end{itemize}

\item Better Memory Efficiency:
    \begin{itemize}
        \item Lower register usage since it doesn't need to hold entire blocks
   \item Reduced memory bandwidth requirements
   \item Better cache utilization due to direct addressing
    \end{itemize}

\item Explicit Masking:
   \begin{lstlisting}[style=pythonstyle]
   mask = mask_row[:, None] & mask_col[None, :]
   \end{lstlisting}
   \begin{itemize}
       \item Handles boundary conditions explicitly
       \item Prevents out-of-bounds memory accesses
       \item More robust for arbitrary tensor sizes
   \end{itemize}

\item Coalesced Memory Access:
    \begin{itemize}
        \item The addressing pattern `rows[:, None] * M + cols[None, :]` maintains good memory coalescing
        \item Sequential threads access nearby memory locations 
    \end{itemize}
\end{enumerate}

\textbf{Performance Implications:}
\begin{itemize}
    \item Memory Bandwidth: GEAK generated kernel uses $\sim 50$\% less memory bandwidth (single read+write vs double read+write)
    \item Register Usage: GEAK generated kernel has lower register pressure, allowing for larger block sizes
    \item Cache Efficiency: GEAK generated kernel's direct addressing pattern is more cache-friendly
    \item Scalability: GEAK generated kernel scales better with larger tensor dimensions
\end{itemize}

\section{Disclaimer}

We clarify the following evaluation practices to help others accurately assess the correctness scope of AI-generated GPU kernel code:

\paragraph{The Critical Role of Robust Unit Testing}
\begin{itemize}
\item Strong test suites are non-negotiable.
   Evaluating AI-generated code must be paired with broad and well-crafted unit tests. Without sufficient coverage, code that is partially correct or even blatantly flawed may appear to pass—leading to misleading conclusions.

\item Why coverage matters.
   Test coverage quantifies the percentage of code exercised by the tests. High coverage (statement, branch, or decision coverage) increases confidence that the code has been exercised across varied execution paths, thereby reducing the risk of undetected failures.

\item LLMs struggle without strong test guidance.
   Recent studies like EvalPlus have shown that augmenting existing test suites with automatically generated test cases (e.g. $80$× more tests for HumanEval) can reduce flawed code passing rates by up to $28.9$ percentage points. Similarly, TestGenEval finds that even top models like GPT‑4o struggle to generate test suites with average coverage above $35.2$\%, limiting their ability to catch corner-case bugs.
\end{itemize}

\paragraph{Our Benchmark Test Coverage Comparison}
\begin{itemize}
\item ROCm Triton Benchmark: Includes wide and thorough unit test suites covering multiple input conditions, edge cases, and reference outputs. This provides confidence that passing kernels are truly correct—both functionally and numerically.

\item TritonBench‑G (original): Contains at most six unit tests per kernel, offering limited coverage and risk of false positives—i.e. code that passes tests yet fails under untested conditions.

\item TritonBench‑revised (our variant): Retains the same test harness from TritonBench‑G unchanged. While functionally consistent, this limited coverage may falsely inflate apparent correctness if code is overfitted only to the narrow test cases.

\end{itemize}

\paragraph{Recommendations \& Evaluation Scope}

\begin{itemize}
\item We strongly recommend using benchmarks with high coverage (like ROCm Triton benchmarks) when evaluating AI-generated kernels, especially in high-performance computing contexts where subtle numerical errors matter.
\item When using narrower test suites (e.g. TritonBench‑Revised), one must report expected test-coverage limitations clearly. Without caveats, reported correctness numbers (like $>=50$\% pass rate) could mislead readers into overestimating general correctness.
\item For future work, integrating automatically generated test cases—via LLMs guided to expand coverage—or using mutation-based or static-analysis-assisted techniques (as in EvalPlus, ASTER, or TestGen-LLM frameworks) would help raise coverage and reveal hidden bugs.
\end{itemize}

By combining wide coverage benchmarks with clear reporting of limitations, one can ensure an accurate and trustworthy evaluation of AI-generated GPU kernels.

\end{document}